\pdfoutput=1
\documentclass[11pt]{article}

\usepackage[final]{acl}

\usepackage{times}
\usepackage{latexsym}

\usepackage{todonotes}
\usepackage{amsmath}
\usepackage{tabularray}
\usepackage{array}
\usepackage{graphicx}
\usepackage{multirow}
\usepackage{ragged2e}
\usepackage{hhline}
\usepackage{arydshln}

\usepackage[T1]{fontenc}
\usepackage[utf8]{inputenc}

\usepackage{microtype}

\usepackage{inconsolata}

\usepackage{cleveref}
\crefname{section}{\S}{\S}
\crefname{table}{Table}{Tables}
\crefname{figure}{Fig.}{Figs.}
\crefname{algorithm}{Alg.}{}
\crefname{ALC@unique}{Line}{Lines}
\crefname{equation}{Eq.}{Eqs.}
\crefname{appendix}{App.}{Apps.}
\crefformat{section}{\S#2#1#3} 

\title{HU at SemEval-2024 Task 8A: Can Contrastive Learning Learn Embeddings to Detect Machine-Generated Text?}

\author{Shubhashis Roy Dipta \\
    Department of CSEE \\ 
    University of Maryland, Baltimore County \\
    Baltimore, Maryland, USA \\
  \texttt{sroydip1@umbc.edu} \\
  \And
  Sadat Shahriar \\
    University of Houston \\
    Houston, Texas, USA \\ 
  \texttt{sshahria@cougarnet.uh.edu} \\
}

\begin{document}

\maketitle

\begin{abstract}
This paper describes our system developed for SemEval-2024 Task 8, ``Multigenerator, Multidomain, and Multilingual Black-Box Machine-Generated Text Detection'' Machine-generated texts have been one of the main concerns due to the use of large language models (LLM) in fake text generation, phishing, cheating in exams, or even plagiarizing copyright materials. A lot of systems have been developed to detect machine-generated text. Nonetheless, the majority of these systems rely on the text-generating model. This limitation is impractical in real-world scenarios, as it's often impossible to know which specific model the user has used for text generation. In this work, we propose a \textbf{single} model based on contrastive learning, which uses \textbf{$\approx$40\% of the baseline's parameters} (149M vs. 355M) but shows a comparable performance on the test dataset (\textit{\textbf{21st out of 137 participants}}). Our key finding is that even without an ensemble of multiple models, a single base model can have comparable performance with the help of data augmentation and contrastive learning. \footnote{Our code is publicly available at \url{https://github.com/dipta007/SemEval24-Task8}}
\end{abstract}

\section{Introduction}
In recent years, Natural Language Processing (NLP) has been totally dependent on Deep Learning rather than statistical machine learning. With multi-task learning \citep{caruana1997multitask}, attention-based transformers \citep{vaswani2017attention}, and the use of Reinforcement Learning in NLP \citep{christiano2017deep}, it has been used in our day-to-day life from mathematical calculations \citep{yang2023gpt} to email writing. But with huge help, it has also been used to generate fake news \citep{Zellers_2019}, to plagiarize copyright materials \citep{dehouche2021plagiarism}, and also to cheat in exams or assignments \citep{cotton2023chatting, fyfe2023cheat}. Humans can identify machine-generated text only at the chance level \citep{Jawahar_2020}. There has been a dire need to develop a system to detect machine-generated text.

Though a lot of works \citep{Badaskar_2008, Gehrmann_2019, Zellers_2019, Jawahar_2020, Ippolito_2020, Chakraborty_2023, Pu_2023, Mitchell_2023, He_2023, Guo_2023} have already been deployed for detecting machine-generated text, with the current development of LLMs, most of the systems are failing to find out which one is human-generated vs. machine-generated (mostly due to the improvement of coherency, fluency and usage of real-world dataset \citep{radford2019language}). In this context, the task "Multigenerator, Multidomain, and Multilingual Black-Box Machine-Generated Text Detection" provides a dataset for training models to classify machine-generated texts. The shared task consists of three sub-tasks: Binary Classification (Machine vs. Human), Multi-class Classification (Which model/human generated this?), and Span Detection (Which part of the text is machine-generated?). A detailed description of the task can be found in the shared task paper \citep{semeval2024task8}. 

In this paper, we describe our final submission on Subtask A (Binary Classification). There were two big challenges of this task: \textbf{First,} five Different models have been used to generate the machine-generated text. \citet{Zellers_2019} has shown that the best defense for machine-generated text is the model itself that was used for generation. However, in reality, there is a massive surge in large language models (LLMs), each with its own unique style of text generation. The challenge in this particular subtask has heightened due to the utilization of five different LLMs. This complexity demands a versatile, model-agnostic architecture capable of detecting text generated by LLMs in a generalized manner. \textbf{Second,} Following the previous challenge, the organizers have employed a different model for generating the validation and test datasets compared to those used in the training set. This implies that the text was drawn from a completely distinct distribution. As a result, participants must develop a generalized model capable of performing effectively regardless of the specific model used in the text generation process.

In response to the key challenges, we have investigated the performance of contrastive learning for this particular task. Contrastive learning has been used as a valuable technique across various domains, including Text Embedding \citep{neelakantan2022text}, Document Embedding \citep{luo2021unsupervised}, Event Embedding \citep{roy-dipta-etal-2023-semantically}, vision \citep{chen2020simple} and Language-Vision learning \citep{radford2021learning}. Notably, unlike the majority of submissions in any shared task like competition, Our final submission utilized a \textbf{single} model to classify the machine-generated texts rather than an ensemble of multiple models. Hence, our contributions to this paper are as follows,

\begin{enumerate}
    \item We proposed a novel data augmentation technique, which nearly makes the data $X$ times bigger ($X$ is the number of models used for data augmentation).
    \item We propose a single unified model that shows a comparable performance on the test dataset.
    \item We have shown that even with a single model, contrastive learning with data augmentation shows a comparable performance, which opens up a door for future exploration.
\end{enumerate}

\section{Related Works}
In this section, we will provide the prior works that have been done in the realm of machine-generated text detection (\cref{rel_fake_text}) and contrastive learning (\cref{rel_fake_text}).

\subsection{Machine Generated Text detection} \label{rel_fake_text}
With the progress of LLMs, much prior research has been done to counter-attack the misuse of the LLMs. Before the attention and transformers, \citet{Badaskar_2008} has shown how the syntactic and semantic features can help in classifying between human and machine-generated text. Later, \citet{Gehrmann_2019} has provided a statistical detection system based on the assumption that the machine samples from the high probability words through max sampling \citep{gu2017trainable}, k-max sampling \citep{fan2018hierarchical}, beam search \citep{shao2017generating}. So, the authors used the probability, rank, and entropy of words as features to classify a machine-generated text. \citet{Jawahar_2020} has shown that state-of-the-art LLM can generate texts with human-like fluency and coherence without grammatical or spelling errors. Lastly, \citet{Mitchell_2023} have used the change of log-probability between the original text and after random perturbation. 

\subsection{Contrastive Learning} \label{rel_con_learn}
Contrastive learning was first introduced in the visual domain \citep{chen2020simple}. Later, it has been widely used in NLP for representation learning \citep{xu2023contrastive, wang2023sncse}, event similarity tasks \citep{gao2022improving} and event modeling \citep{roy-dipta-etal-2023-semantically}. Inspired by the latter works, we have explored whether contrastive learning can help in machine-generated text detection.

\section{System Overview}
Our system is divided into three parts: where the first part is data augmentation (described on \cref{system:data_aug}), the second part is contrastive learning (described on \cref{system:con_learning}), and the last part is the classification head (described on \cref{system:cls}) over the document embeddings.

\subsection{Data Augmentation} \label{system:data_aug}
The dataset provided in the shared task has text and their corresponding label. However, we need a positive and a (hard) negative pair to use contrastive learning. Our main inspiration for using contrastive learning is that as the texts come from two different entities (machine vs. human), the embedding space should also be different. To facilitate the task, we have used a paraphrase model to generate alternate texts for each text in the dataset. In that way, now, every instance of the dataset has one human/machine-generated text and one machine-generated text. We have utilized the human-generated text as the hard negative \footnote{Hard negatives are the total opposite of the given text} and the machine-generated text as the soft positive \footnote{Soft positives expressed the same idea but might not be the exact one}.

Another challenge we faced during the paraphrasing of the dataset is that the texts are long. If we give the whole text to the paraphrase model and ask for alternate text, it gives a much shorter text (an issue we observed in the used paraphrase model). In our primary validation, that gives bad results due to the loss of information while shortening the text. So, instead of giving the whole text at once, we have split the data by end-of-sentence or newline. Then, each sentence was paraphrased on its own and then joined together again to get the previous structure. The technical details behind generating paraphrases and using them for contrastive learning have been discussed in \cref{data_aug} and \cref{pre_trained_enc}, respectively.

\subsection{Contrastive Learning} \label{system:con_learning}
\begin{figure*}[!ht]
    \centering
    \includegraphics[width=\textwidth]{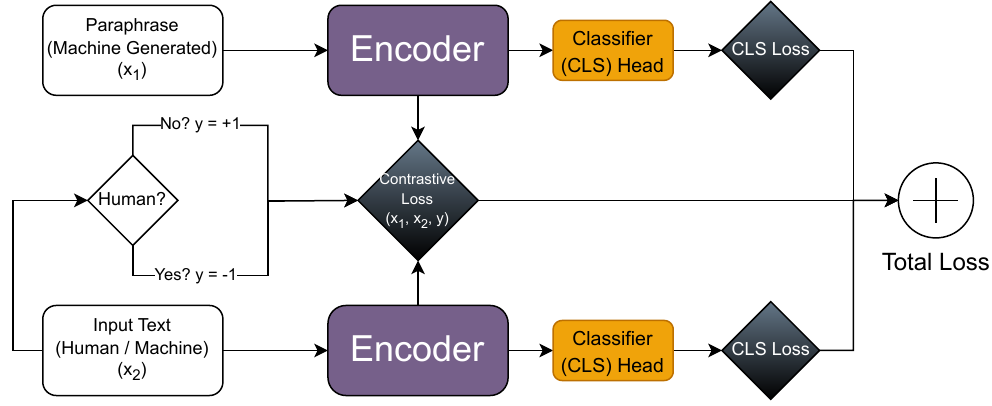}
    \caption{Overview of our model architecture. The same color weights are shared (encoder \& classifier head). Diamond boxes represent the loss function, and the plus sign represents the summation of the three losses. The input to the contrastive loss depends on the original label (y=$+1$ if human, else $-1$).}
    \label{fig:main_model}
\end{figure*}

With the availability of an appropriate dataset for contrastive learning, we proceeded to develop our model. Our main assumption was that the embedding of the machine-generated text and human-generated text would exhibit significant differences. A simple overview of the model is shown in the \cref{fig:main_model}.

The positive and negative data go through the same shared encoder to generate an embedding. This embedding is then used in contrastive learning. We have used the following loss formulation for our contrastive learning:

\begin{equation}
    \begin{aligned}
        \mathcal{L}_{con} &= (1 - y) * cos(x_1, x_2) \\ &+ y * max(0, cos(x_1, x_2)) 
    \end{aligned}
\end{equation}

Here, $x_1$ and $x_2$ are the embeddings of two different pairs, respectively. $cos(x_1, x_2)$ is the cosine-similarity score between two embeddings. $y$ is $+1$ for positive-positive pairs and $-1$ otherwise. In our task, $y$ is $+1$ if the data instance contains text from a machine and the other is paraphrased text and $-1$ if the given text is from a human and the other is paraphrased.

\subsection{Classification Loss} \label{system:cls}
In contrastive learning, our primary objective is to acquire meaningful embeddings containing sufficient information for distinguishing between human-generated and machine-generated text. However, we also need to use a classifier model for the downstream task of outputting the actual label. Keeping that in mind, we have used a simple two-linear layer classifier head on top of the embeddings generated by the encoder. During inference time, we used this classifier head to output the labels. We have optimized our model using a simple binary cross-entropy (BCE) loss.

The total loss of our model is defined as,

\begin{equation}
    \mathcal{L} = \alpha * \mathcal{L}_{con} + \beta * \mathcal{L}_{cls_+} + \gamma * \mathcal{L}_{cls_-}
\end{equation}

Here, $\mathcal{L}_{cls_+}$ is the BCE loss of the positive example, and $\mathcal{L}_{cls_-}$ is the BCE loss of the negative sample of the data instance. $\alpha$, $\beta$, and $\gamma$ are hyperparameters that were set to 0.7, 0.8, and 0.1, respectively, based on validation data.

\section{Experimental Setup}
The following sections are used to describe the technical details behind our data augmentation technique (\cref{data_aug}), Encoder (\cref{pre_trained_enc}), Classifier Head (\cref{cls_head}) and Hyperparameters (\cref{hparams}).

\subsection{Data Augmentation \& Pre-processing} \label{data_aug}
We preprocess the raw input, splitting each document into multiple sentences for paraphrasing. After the preprocessing, we got $\approx$ 3.6 million sentences. Even if we are splitting by new lines or end-of-sentences, we kept exactly the same format during joining, i.e., two new lines rather than 1, to keep most information intact. As the paraphrasing is done on the sentence level rather than the paragraph level, the number of paraphrased sentences is the same as the input sentences (3.6M). So, ideally, we got double the number of training data just by using the data augmentation.

We have tried multiple models from HuggingfaceHub \footnote{\href{https://huggingface.co/ibm/qcpg-sentences}{ibm/qcpg-sentences} \label{footnote:2}} \footnote{\href{https://huggingface.co/ceshine/t5-paraphrase-paws-msrp-opinosis}{ceshine/t5-paraphrase-paws-msrp-opinosis}} to generate paraphrase. In our final submission, we have used \citet{bandel-etal-2022-quality}'s model $^{\ref{footnote:2}}$ for our data augmentation. Use of multiple models or use of prompt-based models \citep{achiam2023gpt, touvron2023llama} for data augmentation has been left out for future exploration due to time and compute constraints. For data split, we use the official train \& dev data split. Only train data is used for data augmentation, and the dev data is used to calculate evaluation metrics.

\subsection{Pre-trained Encoder} \label{pre_trained_enc}
To encode the document, we have used a pre-trained version of longformer-base \citep{beltagy2020longformer} \footnote{\href{https://huggingface.co/jpwahle/longformer-base-plagiarism-detection}{jpwahle/longformer-base-plagiarism-detection}}. The reason behind using this encoder rather than others is, \textbf{One,} longformer is good for getting embeddings for long documents because of using global vs. local attention (more details in \citet{beltagy2020longformer}). \textbf{Second, } the pre-trained version was fine-tuned for paraphrase detection, which is kind of similar to our task.

\subsection{Classifier Head} \label{cls_head}
We have used two linear layers for classifier heads with \textit{tanh} activation loss between them. We also have used a dropout layer between them with a probability of 60\%. The primary rationale for using a high dropout rate was to enhance the model's generalization ability and reduce its dependence on the training data.

\subsection{Hyperparameters} \label{hparams}
For training our model, we have used AdamW \citep{loshchilov2017decoupled} optimizer with a learning rate of 1e-5. We have used a batch size of 2 with gradient accumulation for 8 steps (effective batch size 16). We have used early stopping on the validation data with patience 10. Maximum document length was set to 4096 as most of the documents are large. We use the PyTorch-lightning \footnote{\url{https://lightning.ai/}} library to run the experiments and Weight \& Biases \footnote{\url{https://wandb.ai/}} for logging. All of our experiments are run on NVIDIA Quadro RTX 8000 48GB.

\section{Results}
In this section, we report our results on subtask A and discuss our analysis. Our evaluation is based on the accuracy metric, but we have provided the micro and macro-f1 for better comparison. All the results are averaged on 3 runs with 3 different random seeds.

\subsection{Baseline \& Our Model}
We use the official baseline provided by the task organizers. They have used RoBERTa-large \citep{liu2019roberta} as the encoder and fine-tuned on the train data. Throughout the paper, we refer to this model as $baseline_{rob}$.

We have fine-tuned our model (shown in \cref{fig:main_model}) on the training dataset. Throughout the paper, we refer to this model as $ours_{con}$

In the \cref{tab:results}, we have reported the results on the official test file. $Ours_{con}$ is the final submission, and $Ours_{con}$+ is the modified version of our final model for more analysis (not official results; used for ablation study - details on \cref{sec:ablation}). We can get a comparable result using $~60\%$ fewer parameters than the baseline. In the next section, we will see that after hyperparameter tuning, we can get around 5.7\% improvement over the baseline. This supports our assumption that using a contrastive learning-based method can help machine-generated text identification.

\subsection{Ablation Study} \label{sec:ablation}

\paragraph{Effect of Maximum Sentence Length:} The maximum sentence length is used to tokenize the document. The optimal test score is achieved with a maximum sentence length of 256. This demonstrates that the model can effectively identify machine-generated text even with documents as large as 256 words. This underscores the effectiveness and adaptability of our model's learning capabilities.

\paragraph{Effect of Classification Dropout:} The classification dropout is applied between the two classification layers. Contrary to our initial assumption, the results presented in \cref{tab:results} indicate that using a low dropout rate (as low as 0.0) contributes positively to the model's learning process. This suggests that, even without dropout, the model's generalization to unseen data (text generated by a new model) is enabled primarily through contrastive learning and data augmentation.

\paragraph{Effects of (Effective) Batch Size:} Due to computational constraint, we have used a fixed batch size of 2 and gradient accumulation steps of $\{1, 2, 4, 8, 16, 32, 64\}$ resulting in an effective batch size of $\{2, 4, 8, 16, 32, 64, 128\}$. From the results report on \cref{tab:results}, we found that using only an effective batch size of 2 yielded superior performance compared to gradient accumulation. Notably, this configuration represents the most optimal result obtained following hyperparameter tuning, positioning us at the $8^{th}$ rank in the final standings. This suggests that, in this particular context, the benefits of gradient accumulation may be limited compared to simply using a smaller batch size.

\begin{table}
\centering
\resizebox{\linewidth}{!}{%
\begin{tabular}{l|cccccc}
\multicolumn{1}{l}{} & \begin{tabular}[c]{@{}c@{}}Max Sen\\Length\end{tabular} & \begin{tabular}[c]{@{}c@{}}CLS\\Dropout\end{tabular} & \begin{tabular}[c]{@{}c@{}}Effective\\Batch Size\end{tabular} & Macro-f1 & Micro-f1 & Accuracy \\ 
\hhline{~======}
$Ours_{con}$ & 4096 & 0.6 & 16 & \textbf{88.81} & \textbf{89.07} & \textbf{89.07} \\
$baseline_{rob}$ & - & - & - & - & - & 88.47 \\
 & \multicolumn{6}{c}{Maximum Sentence Length} \\ 
\cdashline{2-7}
$Ours_{con}$+ & 128 & \multirow{6}{*}{0.6} & \multirow{6}{*}{16} & 88.88 & 89.14 & 89.14 \\
$Ours_{con}$+ & 256 &  &  & \textbf{93.30} & \textbf{93.37} & \textbf{93.36} \\
$Ours_{con}$+ & 512 &  &  & 88.78 & 89.04 & 89.04 \\
$Ours_{con}$+ & 1024 &  &  & 90.99 & 91.13 & 91.13 \\
$Ours_{con}$+ & 2048 &  &  & 91.81 & 91.93 & 91.93 \\
$Ours_{con}$ & 4096 &  &  & 88.81 & 89.07 & 89.07 \\
 & \multicolumn{6}{c}{Classification Layer Dropout} \\ 
\cdashline{2-7}
$Ours_{con}$+ & \multirow{5}{*}{4096} & 0 & \multirow{5}{*}{16} & \textbf{92.73} & \textbf{92.81} & \textbf{92.81} \\
$Ours_{con}$+ &  & 0.2 &  & 90.16 & 90.33 & 90.33 \\
$Ours_{con}$+ &  & 0.4 &  & 78.98 & 80.21 & 80.21 \\
$Ours_{con}$ &  & 0.6 &  & 88.81 & 89.07 & 89.07 \\
$Ours_{con}$+ &  & 0.9 &  & 82.60 & 83.31 & 83.31 \\
 & \multicolumn{6}{c}{Effective Batch Size} \\ 
\cdashline{2-7}
$Ours_{con}$+ & \multirow{7}{*}{4096} & \multirow{7}{*}{0.6} & 2 & \textbf{\underline{93.80}} & \textbf{\underline{93.86}} & \textbf{\underline{93.86}} \\
$Ours_{con}$+ &  &  & 4 & 70.79 & 73.52 & 73.52 \\
$Ours_{con}$+ &  &  & 8 & 76.82 & 78.43 & 78.43 \\
$Ours_{con}$ &  &  & 16 & 88.81 & 89.07 & 89.07 \\
$Ours_{con}$+ &  &  & 32 & 79.72 & 80.83 & 80.83 \\
$Ours_{con}$+ &  &  & 64 & 90.64 & 90.80 & 90.80 \\
$Ours_{con}$+ &  &  & 128 & 91.39 & 91.51 & 91.51
\end{tabular}
}
\caption{Macro-f1, Micro-f1, and Accuracy score on the test result. $Ours_{con}$ - final submitted model on the shared task, $baseline_{rob}$ - official baseline model, and $Ours_{con}+$ - modified versions of our final model with more hyperparameter tuning. The \textbf{bold} value signifies the best score within a specific section, whereas the \textbf{\underline{underlined}} value denotes the best score across all sections.}
\label{tab:results}
\end{table}

\section{Conclusion \& Future Work}
In this work, we introduce our contrastive learning-based system, which shows a comparable performance. We demonstrate that a model with half the parameters and without an ensemble of large models or hand-engineered features can show a comparable performance, which requires more exploration in this field. For future work, the use of recent prompt-based models \footnote{\url{https://chat.openai.com/}} can be used for data augmentation. Also, the effect of more advanced contrastive loss, i.e., Triplet loss \citep{chechik2010large} or InfoNCE loss \citep{oord2018representation}, need to be explored.

% \section{Limitations}

\bibliography{custom}

\begin{thebibliography}{38}
\expandafter\ifx\csname natexlab\endcsname\relax\def\natexlab#1{#1}\fi

\bibitem[{Achiam et~al.(2023)Achiam, Adler, Agarwal, Ahmad, Akkaya, Aleman, Almeida, Altenschmidt, Altman, Anadkat et~al.}]{achiam2023gpt}
Josh Achiam, Steven Adler, Sandhini Agarwal, Lama Ahmad, Ilge Akkaya, Florencia~Leoni Aleman, Diogo Almeida, Janko Altenschmidt, Sam Altman, Shyamal Anadkat, et~al. 2023.
\newblock Gpt-4 technical report.
\newblock \emph{arXiv preprint arXiv:2303.08774}.

\bibitem[{Badaskar et~al.(2008)Badaskar, Badaskar, Badaskar, Agarwal, Agarwal, Arora, and Arora}]{Badaskar_2008}
Sameer Badaskar, Sameer Badaskar, Sameer Badaskar, Sonali Agarwal, Sachin Agarwal, Shilpa Arora, and Shilpa Arora. 2008.
\newblock \href {https://doi.org/null} {Identifying real or fake articles: Towards better language modeling}.
\newblock \emph{International Joint Conference on Natural Language Processing}.

\bibitem[{Bandel et~al.(2022)Bandel, Aharonov, Shmueli-Scheuer, Shnayderman, Slonim, and Ein-Dor}]{bandel-etal-2022-quality}
Elron Bandel, Ranit Aharonov, Michal Shmueli-Scheuer, Ilya Shnayderman, Noam Slonim, and Liat Ein-Dor. 2022.
\newblock \href {https://doi.org/10.18653/v1/2022.acl-long.45} {Quality controlled paraphrase generation}.
\newblock In \emph{Proceedings of the 60th Annual Meeting of the Association for Computational Linguistics (Volume 1: Long Papers)}, pages 596--609, Dublin, Ireland. Association for Computational Linguistics.

\bibitem[{Beltagy et~al.(2020)Beltagy, Peters, and Cohan}]{beltagy2020longformer}
Iz~Beltagy, Matthew~E Peters, and Arman Cohan. 2020.
\newblock Longformer: The long-document transformer.
\newblock \emph{arXiv preprint arXiv:2004.05150}.

\bibitem[{Caruana(1997)}]{caruana1997multitask}
Rich Caruana. 1997.
\newblock Multitask learning.
\newblock \emph{Machine learning}, 28:41--75.

\bibitem[{Chakraborty et~al.(2023)Chakraborty, Bedi, Zhu, An, Manocha, and Huang}]{Chakraborty_2023}
Souradip Chakraborty, Amrit~Singh Bedi, Sicheng Zhu, Bang An, Dinesh Manocha, and Furong Huang. 2023.
\newblock \href {https://doi.org/10.48550/arxiv.2304.04736} {On the possibilities of ai-generated text detection}.
\newblock \emph{arXiv.org}.

\bibitem[{Chechik et~al.(2010)Chechik, Sharma, Shalit, and Bengio}]{chechik2010large}
Gal Chechik, Varun Sharma, Uri Shalit, and Samy Bengio. 2010.
\newblock Large scale online learning of image similarity through ranking.
\newblock \emph{Journal of Machine Learning Research}, 11(3).

\bibitem[{Chen et~al.(2020)Chen, Kornblith, Norouzi, and Hinton}]{chen2020simple}
Ting Chen, Simon Kornblith, Mohammad Norouzi, and Geoffrey Hinton. 2020.
\newblock A simple framework for contrastive learning of visual representations.
\newblock In \emph{International conference on machine learning}, pages 1597--1607. PMLR.

\bibitem[{Christiano et~al.(2017)Christiano, Leike, Brown, Martic, Legg, and Amodei}]{christiano2017deep}
Paul~F Christiano, Jan Leike, Tom Brown, Miljan Martic, Shane Legg, and Dario Amodei. 2017.
\newblock Deep reinforcement learning from human preferences.
\newblock \emph{Advances in neural information processing systems}, 30.

\bibitem[{Cotton et~al.(2023)Cotton, Cotton, and Shipway}]{cotton2023chatting}
Debby~RE Cotton, Peter~A Cotton, and J~Reuben Shipway. 2023.
\newblock Chatting and cheating: Ensuring academic integrity in the era of chatgpt.
\newblock \emph{Innovations in Education and Teaching International}, pages 1--12.

\bibitem[{Dehouche(2021)}]{dehouche2021plagiarism}
Nassim Dehouche. 2021.
\newblock Plagiarism in the age of massive generative pre-trained transformers (gpt-3).
\newblock \emph{Ethics in Science and Environmental Politics}, 21:17--23.

\bibitem[{Fan et~al.(2018)Fan, Lewis, and Dauphin}]{fan2018hierarchical}
Angela Fan, Mike Lewis, and Yann Dauphin. 2018.
\newblock Hierarchical neural story generation.
\newblock \emph{arXiv preprint arXiv:1805.04833}.

\bibitem[{Fyfe(2023)}]{fyfe2023cheat}
Paul Fyfe. 2023.
\newblock How to cheat on your final paper: Assigning ai for student writing.
\newblock \emph{AI \& SOCIETY}, 38(4):1395--1405.

\bibitem[{Gao et~al.(2022)Gao, Wang, Yu, Zhao, Ng, and Xu}]{gao2022improving}
Jun Gao, Wei Wang, Changlong Yu, Huan Zhao, Wilfred Ng, and Ruifeng Xu. 2022.
\newblock Improving event representation via simultaneous weakly supervised contrastive learning and clustering.
\newblock \emph{arXiv preprint arXiv:2203.07633}.

\bibitem[{Gehrmann et~al.(2019)Gehrmann, Gehrmann, Strobelt, Strobelt, Rushton, Rush, Rush, and Rush}]{Gehrmann_2019}
Sebastian Gehrmann, Sebastian Gehrmann, Hendrik Strobelt, Hendrik Strobelt, Gérard Rushton, Alexander~M. Rush, Alexander~M. Rush, and Alexander~M. Rush. 2019.
\newblock \href {https://doi.org/10.18653/v1/p19-3019} {Gltr: Statistical detection and visualization of generated text}.
\newblock \emph{Annual Meeting of the Association for Computational Linguistics}.

\bibitem[{Gu et~al.(2017)Gu, Cho, and Li}]{gu2017trainable}
Jiatao Gu, Kyunghyun Cho, and Victor~OK Li. 2017.
\newblock Trainable greedy decoding for neural machine translation.
\newblock \emph{arXiv preprint arXiv:1702.02429}.

\bibitem[{Guo et~al.(2023)Guo, Zhang, Wang, Jiang, Nie, Ding, Yue, and Wu}]{Guo_2023}
Biyang Guo, Xin Zhang, Ziyuan Wang, Minqi Jiang, Jinran Nie, Yuxuan Ding, Jianwei Yue, and Yupeng Wu. 2023.
\newblock \href {https://doi.org/10.48550/arxiv.2301.07597} {How close is chatgpt to human experts? comparison corpus, evaluation, and detection}.
\newblock \emph{arXiv.org}.

\bibitem[{He et~al.(2023)He, Shen, Chen, Backes, and Zhang}]{He_2023}
Xiaotong He, Xinyue Shen, Zeyuan Chen, Michael Backes, and Yang Zhang. 2023.
\newblock \href {https://doi.org/10.48550/arxiv.2303.14822} {Mgtbench: Benchmarking machine-generated text detection}.
\newblock \emph{arXiv.org}.

\bibitem[{Ippolito et~al.(2020)Ippolito, Ippolito, Duckworth, Duckworth, Callison-Burch, Callison-Burch, Eck, and Eck}]{Ippolito_2020}
Daphne Ippolito, Daphne Ippolito, Daniel Duckworth, Daniel Duckworth, Chris Callison-Burch, Chris Callison-Burch, Douglas Eck, and Douglas Eck. 2020.
\newblock \href {https://doi.org/10.18653/v1/2020.acl-main.164} {Automatic detection of generated text is easiest when humans are fooled}.
\newblock \emph{Annual Meeting of the Association for Computational Linguistics}.

\bibitem[{Jawahar et~al.(2020)Jawahar, Jawahar, Abdul-Mageed, Abdul-Mageed, Lakshmanan, and Lakshmanan}]{Jawahar_2020}
Ganesh Jawahar, Ganesh Jawahar, Muhammad Abdul-Mageed, Muhammad Abdul-Mageed, Laks V.~S. Lakshmanan, and Laks V.~S. Lakshmanan. 2020.
\newblock \href {https://doi.org/10.18653/v1/2020.coling-main.208} {Automatic detection of machine generated text: A critical survey.}
\newblock \emph{International Conference on Computational Linguistics}.

\bibitem[{Liu et~al.(2019)Liu, Ott, Goyal, Du, Joshi, Chen, Levy, Lewis, Zettlemoyer, and Stoyanov}]{liu2019roberta}
Yinhan Liu, Myle Ott, Naman Goyal, Jingfei Du, Mandar Joshi, Danqi Chen, Omer Levy, Mike Lewis, Luke Zettlemoyer, and Veselin Stoyanov. 2019.
\newblock Roberta: A robustly optimized bert pretraining approach.
\newblock \emph{arXiv preprint arXiv:1907.11692}.

\bibitem[{Loshchilov and Hutter(2017)}]{loshchilov2017decoupled}
Ilya Loshchilov and Frank Hutter. 2017.
\newblock Decoupled weight decay regularization.
\newblock \emph{arXiv preprint arXiv:1711.05101}.

\bibitem[{Luo et~al.(2021)Luo, Cheng, Ni, Yu, Zhang, Zong, Liu, Chen, Song, Chen et~al.}]{luo2021unsupervised}
Dongsheng Luo, Wei Cheng, Jingchao Ni, Wenchao Yu, Xuchao Zhang, Bo~Zong, Yanchi Liu, Zhengzhang Chen, Dongjin Song, Haifeng Chen, et~al. 2021.
\newblock Unsupervised document embedding via contrastive augmentation.
\newblock \emph{arXiv preprint arXiv:2103.14542}.

\bibitem[{Mitchell et~al.(2023)Mitchell, Lee, Khazatsky, Manning, and Finn}]{Mitchell_2023}
Eric Mitchell, Yoonho Lee, Alexander Khazatsky, Christopher~D. Manning, and Chelsea Finn. 2023.
\newblock \href {https://doi.org/10.48550/arxiv.2301.11305} {Detectgpt: Zero-shot machine-generated text detection using probability curvature}.
\newblock \emph{International Conference on Machine Learning}.

\bibitem[{Neelakantan et~al.(2022)Neelakantan, Xu, Puri, Radford, Han, Tworek, Yuan, Tezak, Kim, Hallacy et~al.}]{neelakantan2022text}
Arvind Neelakantan, Tao Xu, Raul Puri, Alec Radford, Jesse~Michael Han, Jerry Tworek, Qiming Yuan, Nikolas Tezak, Jong~Wook Kim, Chris Hallacy, et~al. 2022.
\newblock Text and code embeddings by contrastive pre-training.
\newblock \emph{arXiv preprint arXiv:2201.10005}.

\bibitem[{Oord et~al.(2018)Oord, Li, and Vinyals}]{oord2018representation}
Aaron van~den Oord, Yazhe Li, and Oriol Vinyals. 2018.
\newblock Representation learning with contrastive predictive coding.
\newblock \emph{arXiv preprint arXiv:1807.03748}.

\bibitem[{Pu et~al.(2023)Pu, Sarwar, Abdullah, Rehman, Kim, Bhattacharya, Javed, and Viswanath}]{Pu_2023}
Jiameng Pu, Zain Sarwar, Sifat~Muhammad Abdullah, Abdullah Rehman, Yoonjin Kim, Parantapa Bhattacharya, Mobin Javed, and Bimal Viswanath. 2023.
\newblock \href {https://doi.org/10.1109/sp46215.2023.10179387} {Deepfake text detection: Limitations and opportunities}.
\newblock \emph{IEEE Symposium on Security and Privacy}.

\bibitem[{Radford et~al.(2021)Radford, Kim, Hallacy, Ramesh, Goh, Agarwal, Sastry, Askell, Mishkin, Clark et~al.}]{radford2021learning}
Alec Radford, Jong~Wook Kim, Chris Hallacy, Aditya Ramesh, Gabriel Goh, Sandhini Agarwal, Girish Sastry, Amanda Askell, Pamela Mishkin, Jack Clark, et~al. 2021.
\newblock Learning transferable visual models from natural language supervision.
\newblock In \emph{International conference on machine learning}, pages 8748--8763. PMLR.

\bibitem[{Radford et~al.(2019)Radford, Wu, Child, Luan, Amodei, Sutskever et~al.}]{radford2019language}
Alec Radford, Jeffrey Wu, Rewon Child, David Luan, Dario Amodei, Ilya Sutskever, et~al. 2019.
\newblock Language models are unsupervised multitask learners.
\newblock \emph{OpenAI blog}, 1(8):9.

\bibitem[{Roy~Dipta et~al.(2023)Roy~Dipta, Rezaee, and Ferraro}]{roy-dipta-etal-2023-semantically}
Shubhashis Roy~Dipta, Mehdi Rezaee, and Francis Ferraro. 2023.
\newblock \href {https://doi.org/10.18653/v1/2023.starsem-1.31} {Semantically-informed hierarchical event modeling}.
\newblock In \emph{Proceedings of the 12th Joint Conference on Lexical and Computational Semantics (*SEM 2023)}, pages 353--369, Toronto, Canada. Association for Computational Linguistics.

\bibitem[{Shao et~al.(2017)Shao, Gouws, Britz, Goldie, Strope, and Kurzweil}]{shao2017generating}
Louis Shao, Stephan Gouws, Denny Britz, Anna Goldie, Brian Strope, and Ray Kurzweil. 2017.
\newblock Generating high-quality and informative conversation responses with sequence-to-sequence models.
\newblock \emph{arXiv preprint arXiv:1701.03185}.

\bibitem[{Touvron et~al.(2023)Touvron, Lavril, Izacard, Martinet, Lachaux, Lacroix, Rozi{\`e}re, Goyal, Hambro, Azhar et~al.}]{touvron2023llama}
Hugo Touvron, Thibaut Lavril, Gautier Izacard, Xavier Martinet, Marie-Anne Lachaux, Timoth{\'e}e Lacroix, Baptiste Rozi{\`e}re, Naman Goyal, Eric Hambro, Faisal Azhar, et~al. 2023.
\newblock Llama: Open and efficient foundation language models.
\newblock \emph{arXiv preprint arXiv:2302.13971}.

\bibitem[{Vaswani et~al.(2017)Vaswani, Shazeer, Parmar, Uszkoreit, Jones, Gomez, Kaiser, and Polosukhin}]{vaswani2017attention}
Ashish Vaswani, Noam Shazeer, Niki Parmar, Jakob Uszkoreit, Llion Jones, Aidan~N Gomez, {\L}ukasz Kaiser, and Illia Polosukhin. 2017.
\newblock Attention is all you need.
\newblock \emph{Advances in neural information processing systems}, 30.

\bibitem[{Wang and Dou(2023)}]{wang2023sncse}
Hao Wang and Yong Dou. 2023.
\newblock Sncse: contrastive learning for unsupervised sentence embedding with soft negative samples.
\newblock In \emph{International Conference on Intelligent Computing}, pages 419--431. Springer.

\bibitem[{Wang et~al.(2024)Wang, Mansurov, Ivanov, Su, Shelmanov, Tsvigun, Whitehouse, Afzal, Mahmoud, Puccetti, Arnold, Aji, Habash, Gurevych, and Nakov}]{semeval2024task8}
Yuxia Wang, Jonibek Mansurov, Petar Ivanov, Jinyan Su, Artem Shelmanov, Akim Tsvigun, Chenxi Whitehouse, Osama~Mohammed Afzal, Tarek Mahmoud, Giovanni Puccetti, Thomas Arnold, Alham~Fikri Aji, Nizar Habash, Iryna Gurevych, and Preslav Nakov. 2024.
\newblock {SemEval}-2024 task 8: Multigenerator, multidomain, and multilingual black-box machine-generated text detection.
\newblock In \emph{Proceedings of the 18th International Workshop on Semantic Evaluation}, SemEval 2024, Mexico City, Mexico.

\bibitem[{Xu et~al.(2023)Xu, Xie, Li, Wang, Wang, and Li}]{xu2023contrastive}
Lingling Xu, Haoran Xie, Zongxi Li, Fu~Lee Wang, Weiming Wang, and Qing Li. 2023.
\newblock Contrastive learning models for sentence representations.
\newblock \emph{ACM Transactions on Intelligent Systems and Technology}, 14(4):1--34.

\bibitem[{Yang et~al.(2023)Yang, Ding, Lv, Jiang, He, Guo, Bai, and Tang}]{yang2023gpt}
Zhen Yang, Ming Ding, Qingsong Lv, Zhihuan Jiang, Zehai He, Yuyi Guo, Jinfeng Bai, and Jie Tang. 2023.
\newblock Gpt can solve mathematical problems without a calculator.
\newblock \emph{arXiv preprint arXiv:2309.03241}.

\bibitem[{Zellers et~al.(2019)Zellers, Zellers, Holtzman, Holtzman, Rashkin, Rashkin, Bisk, Bisk, Farhadi, Farhadi, Roesner, Roesner, Choi, and Choi}]{Zellers_2019}
Rowan Zellers, Rowan Zellers, Ari Holtzman, Ari Holtzman, Hannah Rashkin, Hannah Rashkin, Yonatan Bisk, Yonatan Bisk, Ali Farhadi, Ali Farhadi, Franziska Roesner, Franziska Roesner, Yejin Choi, and Yejin Choi. 2019.
\newblock \href {https://doi.org/null} {Defending against neural fake news}.
\newblock \emph{Neural Information Processing Systems}.

\end{thebibliography}

\appendix

% \section{Example Appendix}
% \label{sec:appendix}

% This is an appendix.

\end{document}